\documentclass[letterpaper, 10 pt, conference]{ieeeconf}  

\IEEEoverridecommandlockouts                              

\usepackage{epsfig}
\usepackage{times}
\usepackage{amsmath}
\usepackage{amssymb}
\usepackage{graphicx}
\usepackage{cite}
\usepackage{subcaption}
\usepackage[]{algorithm2e}
\usepackage{framed}

\usepackage{multirow} 
\usepackage{booktabs} 

\title{\LARGE \bf
An Insect-scale Self-sufficient Rolling Microrobot
}

\author{Palak Bhushan$^{*}$ and Claire Tomlin$^{*}$
\thanks{$^{*}$The authors are with the Department of EECS, University of California, Berkeley, CA 94720, USA.
        {\tt\small palak@berkeley.edu} \text{(corresponding author)}, {\tt\small tomlin@berkeley.edu}.}%
}

\begin{document}

\maketitle
\thispagestyle{empty}
\pagestyle{empty}

\begin{abstract}
We design an insect-sized rolling microrobot driven by continuously rotating wheels. It measures 18mm $\times$ 8mm $\times$ 8mm. There are 2 versions of the robot - a 96mg laser-powered one and a 130mg supercapacitor powered one. The robot can move at 27mm/s (1.5 body lengths per second) with wheels rotating at 300$^\circ$/s, while consuming an average power of 2.5mW. Neither version has any electrical wires coming out of it, with the supercapacitor powered robot also being self-sufficient and is able to roll freely for 8 seconds after a single charge. Low-voltage electromagnetic actuators (1V-3V) along with a novel double-ratcheting mechanism enable the operation of this device. It is, to the best of our knowledge, the lightest and fastest self-sufficient rolling microrobot reported yet. 
\end{abstract}

\section{Introduction}

Among milligram-scale microrobots (or, microbots), the flying kind are well heard of due to their visually appealing flapping wing kinematics, together with the inherent difficulty in making these owing to the high power demand of flight \cite{wood_liftoff, zhang16, baybug18, baybug19}. Thus insect-scale flying microbots are all tethered with the exception of \cite{robofly18} but even that takes off just for a split second on laser power before falling to the ground. 

The design requirements for ground-based microbots however are much relaxed compared to those of flying ones \cite{actuator_selection}, in part due to  
the fact that they are not required to lift their own weight. 
Yet we have seen few 100mg-scale robots that are self-sufficient \cite{pister_10mg}. Most of the electrical-powered designs are tethered \cite{contreras17, saito16}, due to the high-voltage, high-current, and/or, high-power demands on the drive electronics and the power source. 

There have been prior works that are untethered but these mostly require a controlled environment, like a changing external field \cite{bergbreiter_1mg} or an electrical grid surface \cite{rus06} to function, restricting their global operation. Recently a self-sufficient bot was reported \cite{lin_crawling} weighting 200mg with a supercapacitor as its power source. It crawls at 2mm/s just like a bristlebot, that is, using anisotropic forward versus backward friction coefficient between its legs and the ground. But this makes the bot's motion very sensitive to the surface properties, with rougher surfaces potentially rendering it useless. 

\begin{figure}
\centering
\epsfig{file=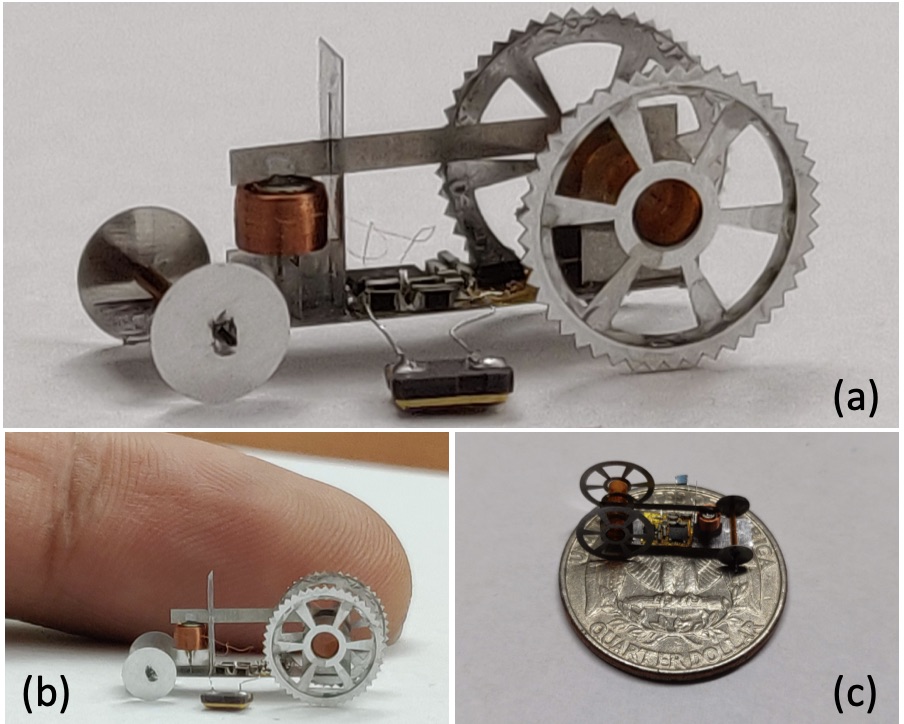,width=3.4in}
\vspace{-1.7em}
\caption{\small{(a) Supercapacitor powered rolling microbot. Compared to (b) an index finger, and, (c) a quarter dollar.}}
\vspace{-1em}
\label{fig:main-pic}
\end{figure}

Rolling and walking locomotions are more robust than bristlebots to changes in the environment's surface, but producing these motions requires generating a continuously rotating motion in contrast to the small-displacement oscillatory actuators available at milligram scales. So we design a new double-ratchet mechanism that converts small periodic motions to large continuous rotations by anisotropically adding up the small motions. The principle behind is similar to some other designs like the inchworm motor \cite{inchworm02, inchworm12} which converts tiny motions of an actuator to large motions of a shuttle. Note that we still use anisotropy in our mechanism, but it has been shifted from the environment to inside the mechanism which we can fully control. The double-ratchet constructed here turns only on clockwise inputs. 

In order to simplify the drive electronics, we take a low-voltage actuation route by using electromagnetic actuation, that is, a magnet plus coil system. Low-voltage approach has previously been taken with flying microbots \cite{zhang16, baybug18, baybug19} but not with ground based ones. 
The use of a double-ratchet to keep the magnet displacements low ensures higher average magnetic fields seen by the coil. Use of the mechanical advantage principle to keep the actuation forces low ensures low current in the coil. Both these strategies further simplify the power electronics by lowering the current and power demands. 

Low-voltage and low-current requirements enable the use of onboard low-power light-weight power sources. The laser-powered version of the bot can operate indefinitely using a 1mg photovoltaic (PV) cell, but the laser needs to be pointed accurately on to the PV cell which becomes a challenge when the bot is moving fast. Thus to demonstrate an untethered motion that is not intermittent we use an onboard 24mg supercapacitor that can power the bot for 8 seconds. 

\section{Design}

\subsection{Micro-ratchet} 
\begin{figure}[h]
\vspace{-0.6em}
\centering
\epsfig{file=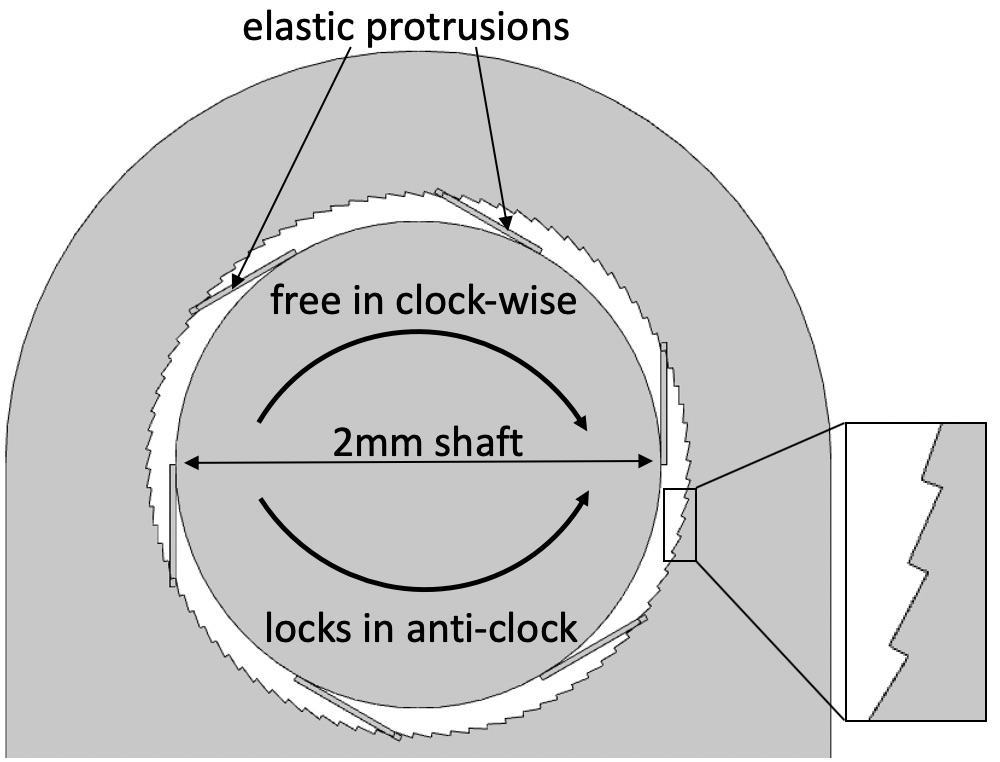,width=2.8in}
\vspace{-0.4em}
\caption{\small{Cross-section of a micro-ratchet mechanism made using flexible beams on a shaft and a patterned hole. The peaks in the pattern are 25$\mu$m high and are spaced 4$^\circ$, or, approximately 70$\mu$m apart. }}
\vspace{-0.5em}
\label{fig:ratchet}
\end{figure}

The basic building block of our mechanism is a ratchet with its cross-section portrayed in Figure \ref{fig:ratchet}. The inner shaft is free to rotate relative to the outer ring when it is rotated in a clockwise direction relative to the ring. Under this operating condition, the elastic protrusions emanating from the shaft slide over the zig-zag patterns on the inner perimeter of the ring. These elastic beams are bent by 25$\mu$m more (in addition to any pre-deflection) when encountering the peaks in the pattern. When rotated anti-clockwise, the shaft locks relative to the ring. In this reverse operation, the elastic beams push the falling edge of the pattern head-on, and motion can only be achieved if the beams buckle. This buckling requires orders of magnitude higher torque compared to the simple sliding motion from before and this configuration can be considered as locked for the purposes of this paper. 

\begin{figure}[h]
\centering
\vspace{-1em}
\epsfig{file=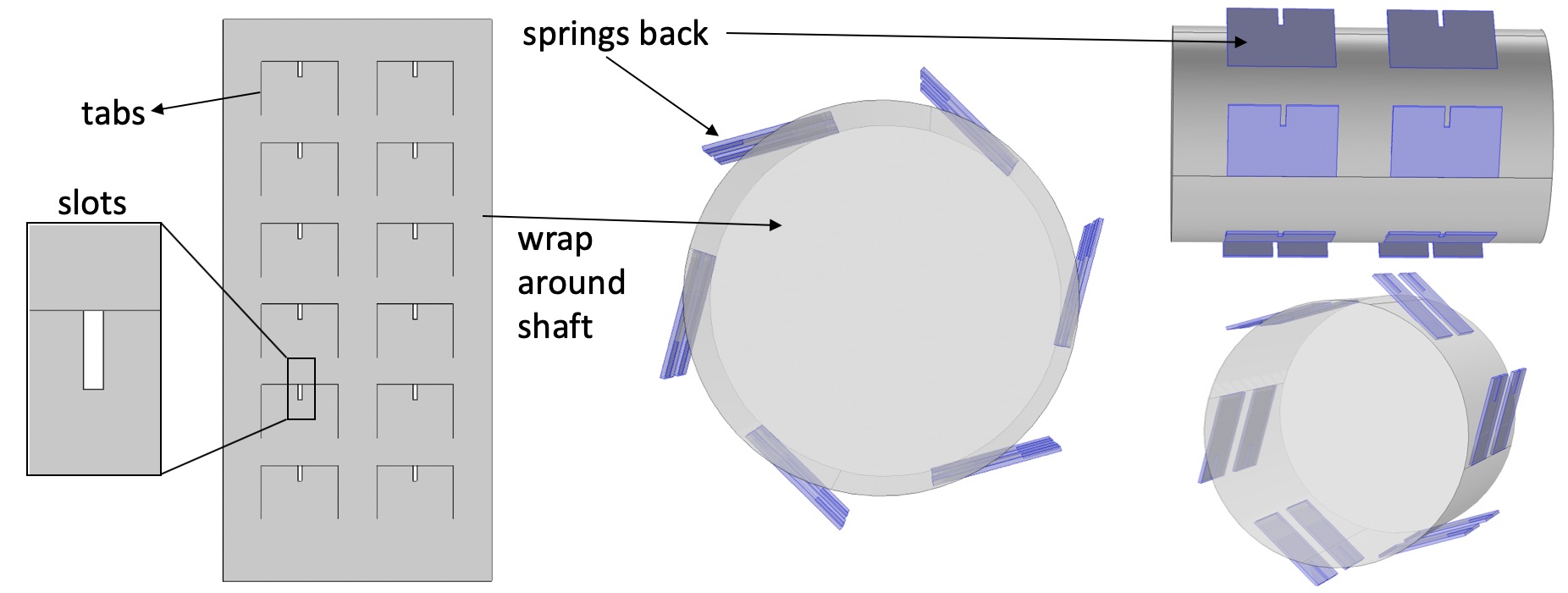,width=3.4in}
\vspace{-1.7em}
\caption{\small{Fabrication of the shaft for the micro-ratchet mechanism. 60$^\circ$ spaced flexible beams are obtained by wrapping a laser cut Kapton sheet with tabs on to a Kapton tube.}}
\vspace{-0.5em}
\label{fig:shaft}
\end{figure}
Figure \ref{fig:shaft} shows the construction of the inner shaft. 12 tabs are laser-cut on a 12.7$\mu$m-thick Kapton sheet. This laser-cut sheet is then rolled on to a 2mm-diameter Kapton tube and glued in place. The rest of the sheet adheres to the curved surface of the tube due to the glue, but the unglued tabs retain their planar shape thus acting as our desired elastic protrusions. The fabricated shaft can be seen in Figure \ref{fig:real-ratchet}. Rectangular slots cut into each of the 12 tabs will help in keeping the outer ring in place as will be seen next. 

\begin{figure}[ht]
\centering
\epsfig{file=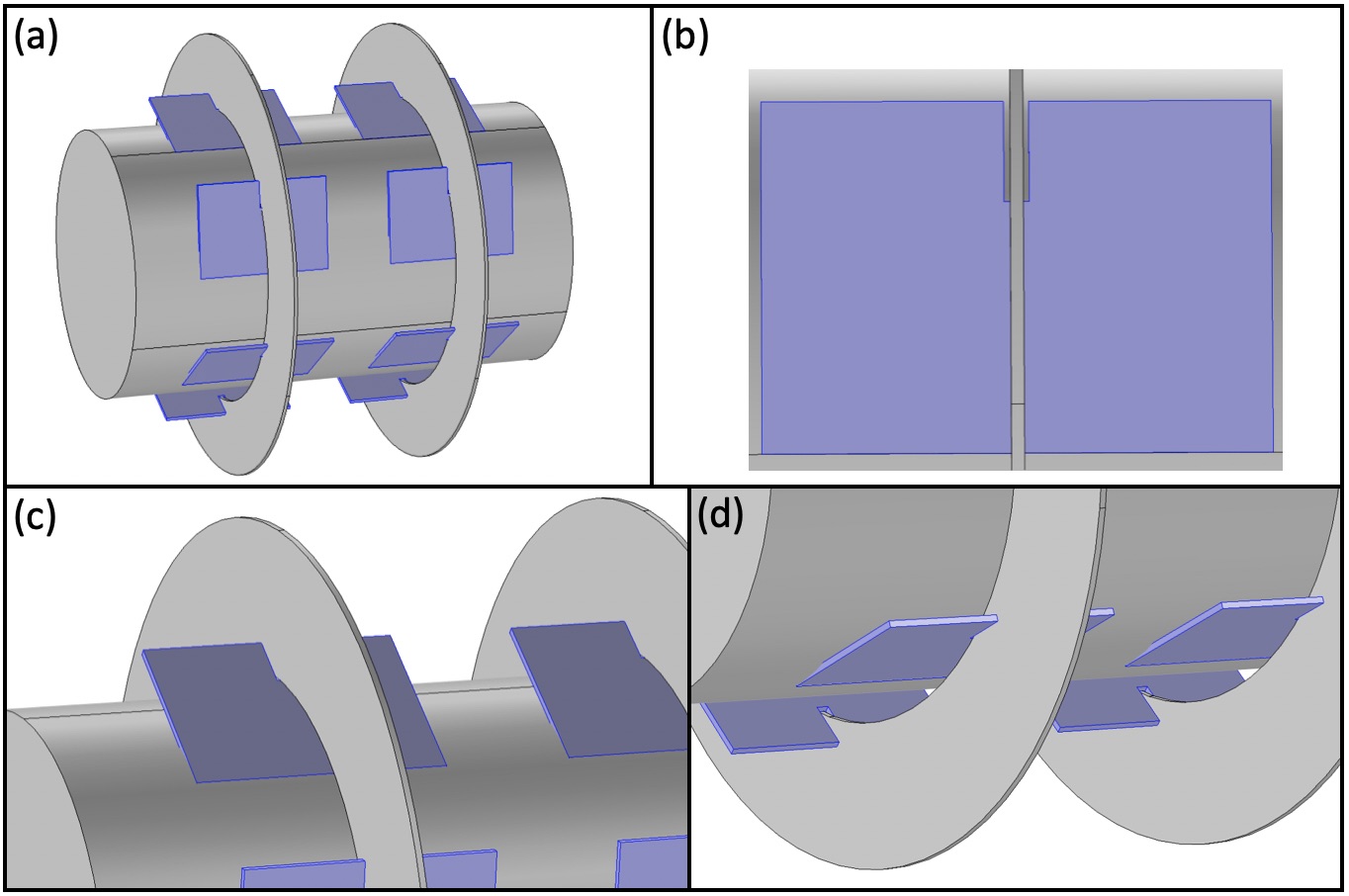,width=3.4in}
\vspace{-1.7em}
\caption{\small{(a) Laser-cut patterned steel rings are slid in to the shaft such that (b) the ring passes through the slots in each of the tabs/elastic beams. (c) \& (d) show a better view of the rings passing through the slots. }}
\vspace{-1.5em}
\label{fig:ring-shaft}
\end{figure}
Rings with patterned holes are laser cut using 25$\mu$m-thick stainless steel. These rings slide into the slots previously cut in each of the tabs as seen in Figure \ref{fig:ring-shaft}. The slots prohibit any sideways motion of the rings, but there is still a slight `give' due to the clearances between the slot and ring, that is, due to the ring being thinner than the slot width. This play can cause the rings to not be perpendicular to the shaft. Thus, in each ratchet a set of 2 rings is used in conjunction to reduce the play and avoid any parasitic motion. The rings are joined using 3 carbon fibre (CF) rods placed 120$^\circ$ apart as seen in Figure \ref{fig:double-ratchet}(a). 

\subsection{Backlash} 
Backlash is the maximum amount the shaft can rotate in the anti-clockwise direction before locking to the ring. From Figure  \ref{fig:ratchet} one can notice that the elastic beams would only need to slide a maximum amount equal to the peak separation in the patterned hole before hitting a falling edge. In reality, this number is even smaller. The elastic beams are not spaced apart at exactly 60$^\circ$ relative to each other due to assembly imperfections, and hence the contact points of the 6 beams are uniformly distributed over the peak separation. Thus, the backlash is estimated to be $1/6^{\texttt{th}}$ of the peak separation, that is, $4^\circ/6 \approx 0.67^\circ$. 

\subsection{Double-ratchet} 

\begin{figure}[h]
\centering
\epsfig{file=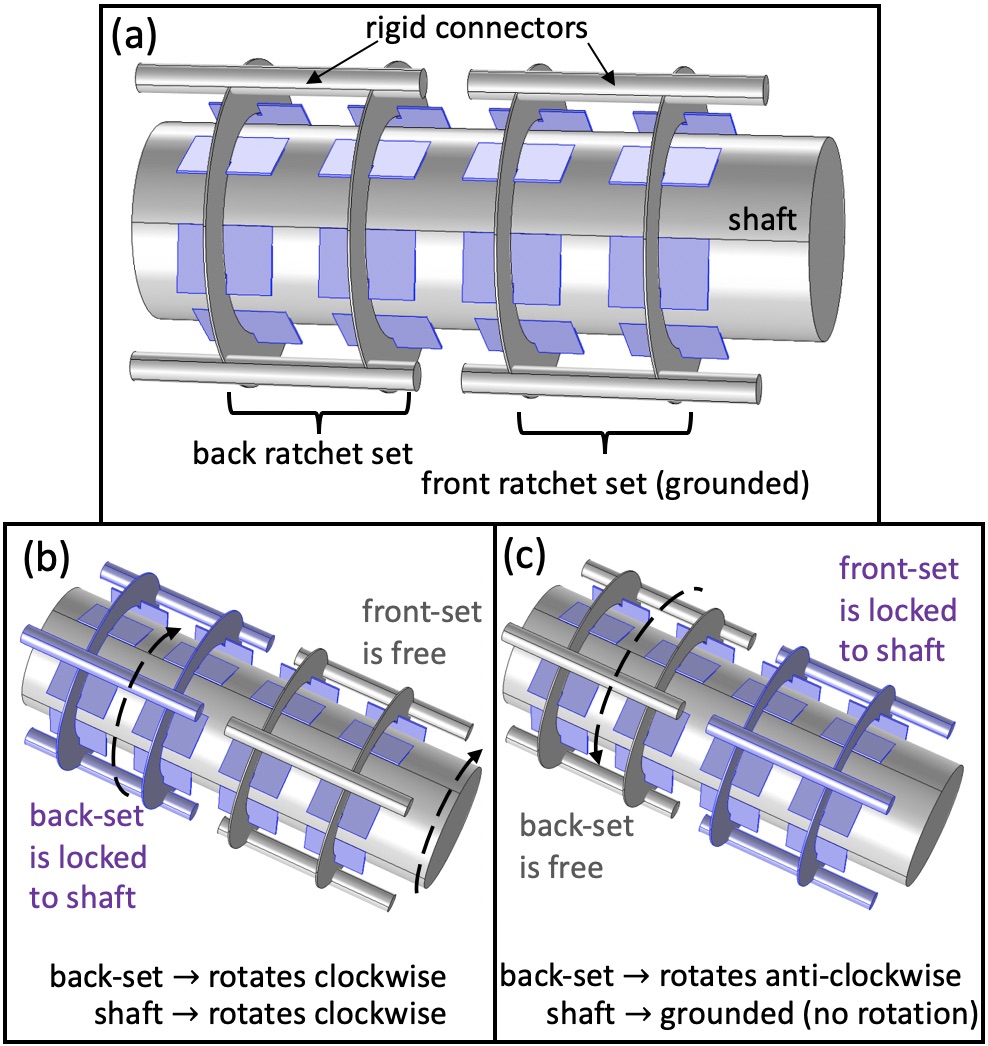,width=3.4in}
\vspace{-1.7em}
\caption{\small{Operation principle of the double ratchet.}}
\vspace{-0.5em}
\label{fig:double-ratchet}
\end{figure}
\begin{figure}[h]
\centering
\epsfig{file=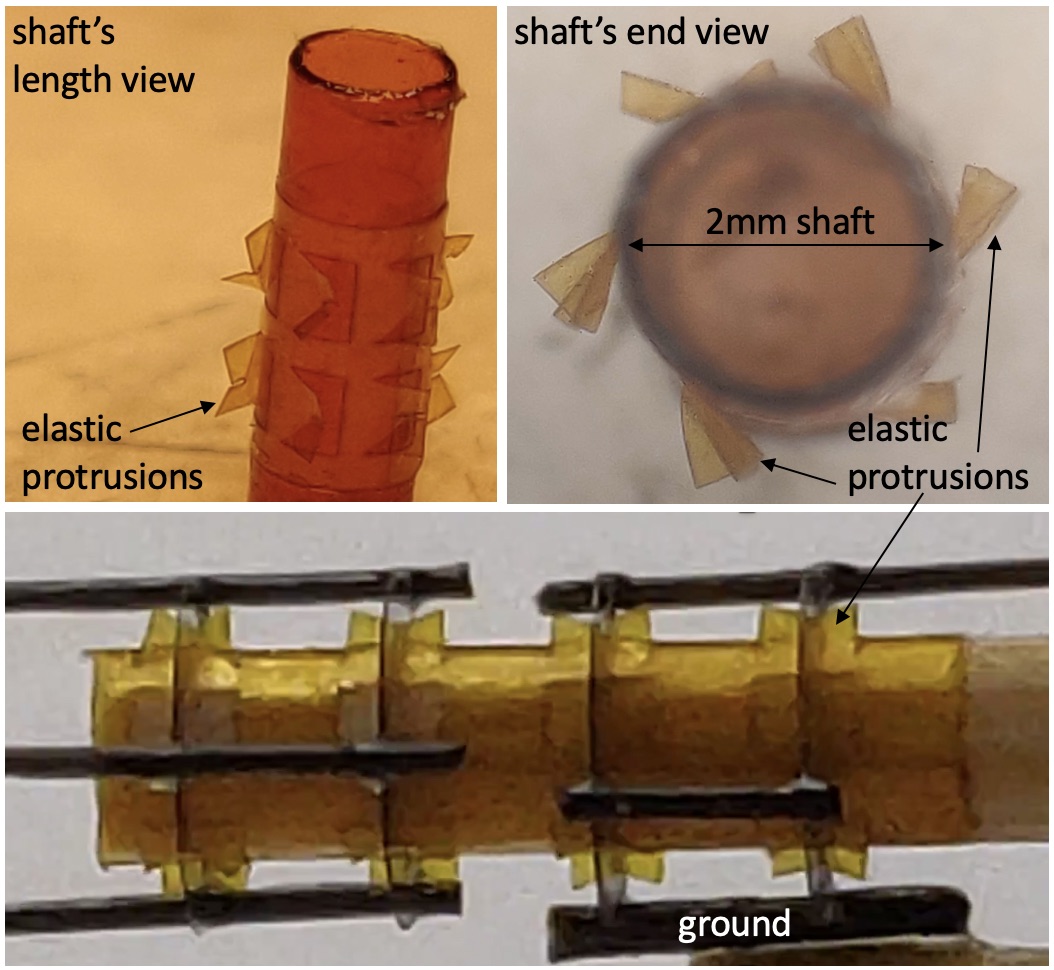,width=3.4in}
\vspace{-1.7em}
\caption{\small{Fabricated double ratchet corresponding to Fig. \ref{fig:double-ratchet}. Seen in black are the 0.3mm CF rods that join the pairs of rings. }}
\vspace{-1em}
\label{fig:real-ratchet}
\end{figure}

Now we seek a mechanism whose output rectifies and adds up the provided input motions. 
Two ratchets are connected together via a common shaft as shown in Figure \ref{fig:double-ratchet}. The front ratchet's rings are grounded. Input is provided at the back ratchet's rings, and the common shaft acts as the output. When the input is rotated clockwise, the back ratchet locks to the shaft but the front ratchet is free to rotate relative to the shaft. Thus, the shaft rotates clockwise. When the input is rotated anti-clockwise, the back ratchet is free to rotate relative to the shaft but the front ratchet locks to the shaft and prohibits it from rotating anti-clockwise. Thus, the shaft remains stationary. Providing periodic clockwise plus anti-clockwise motion at the input results in the shaft adding adding up all the clockwise motions and neglecting any anti-clockwise motions. 
The fabricated double-ratchet assembly can be seen in Figure \ref{fig:real-ratchet}. 

\subsection{Electromagnetic Actuator} 
\begin{figure}[h]
\centering
\epsfig{file=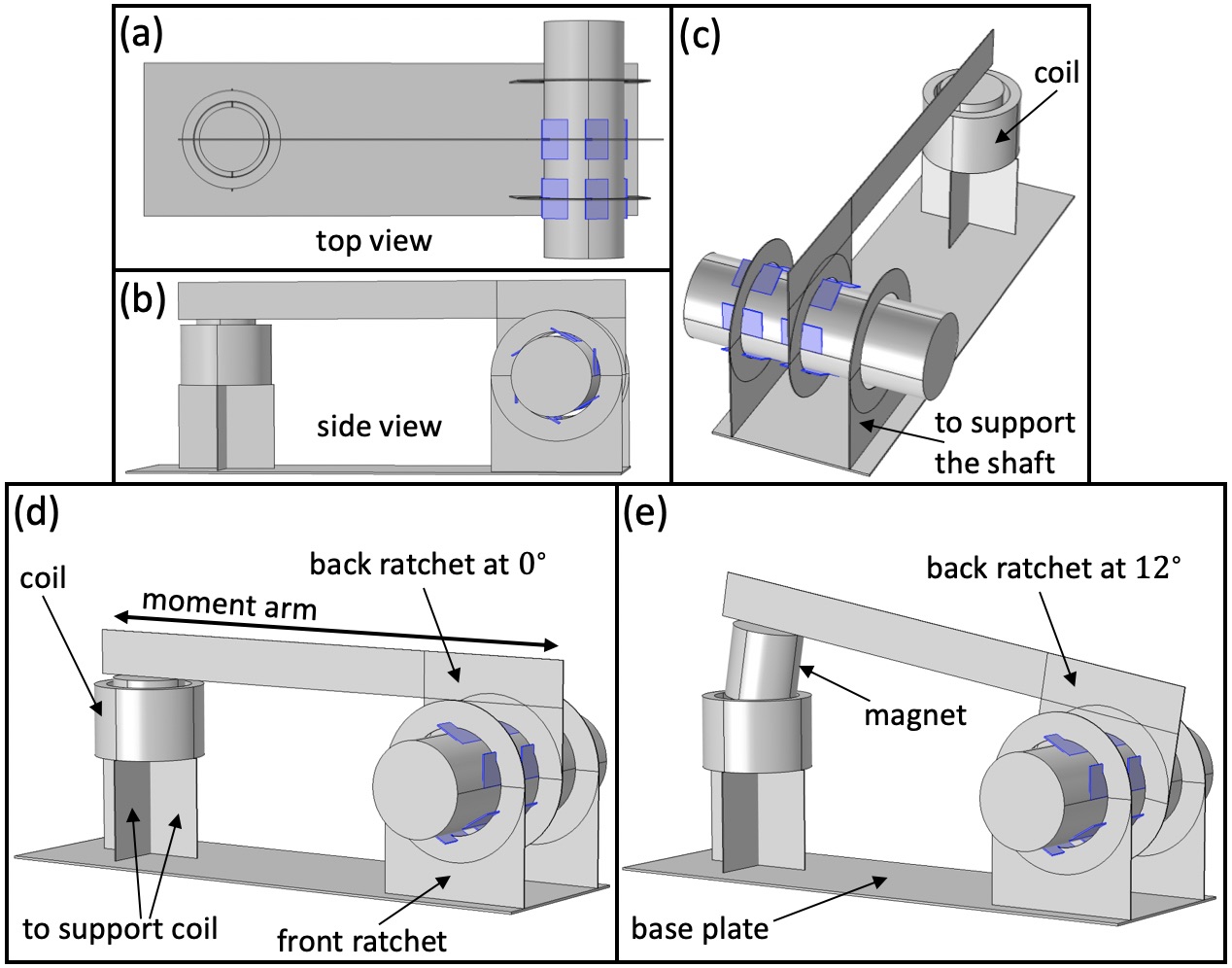,width=3.4in}
\vspace{-1.7em}
\caption{\small{An electromagnetic actuator (magnet + coil) driving the input ratchet via a long moment arm. Rings and coil supports are attached to a common base plate that acts as a mechanical ground.}}
\vspace{-1.5em}
\label{fig:actuator}
\end{figure}

Here, instead of using a set of 2 rings for each of the front and back ratchets, we now use a single ring for both (see Figure \ref{fig:actuator}). This is done to reduce the footprint of the device. We ensure the perpendicularity of the rings and the shaft, which is now lost due to the singular use of the rings, by adding new constraints. Perpendicularity of the grounded ring to the shaft is maintained by balancing the shaft using a third non-patterned ring. Perpendicularity of the input ring to the shaft is maintained by restricting the motion plane of the input. 

The coil is custom made from a 12$\mu$m-thin Copper wire which is array wound $n_\text{turns} = 96\times16$ number of times. It has an inner diameter of 1.9mm, an outer diameter of 2.45mm, and a height of 1.6mm, with resistance $\approx 1500\Omega$. The NdFeB magnet is of grade N52 with 1.6mm diameter and height. The magnet is attached to the input arm of the double-ratchet mechanism. The grounded ring is attached to the rectangular base plate made from 50$\mu$m-thick Aluminum sheet. The shaft is further supported by a non-patterned ring to ensure its perpendicularity. The input ring is attached via a long moment arm to the magnet which is concentric to the coil in its rest state. The fully deflected position of the magnet is chosen such that one of its pole faces is still almost inside the coil. 

\begin{figure}[h]
\centering
\epsfig{file=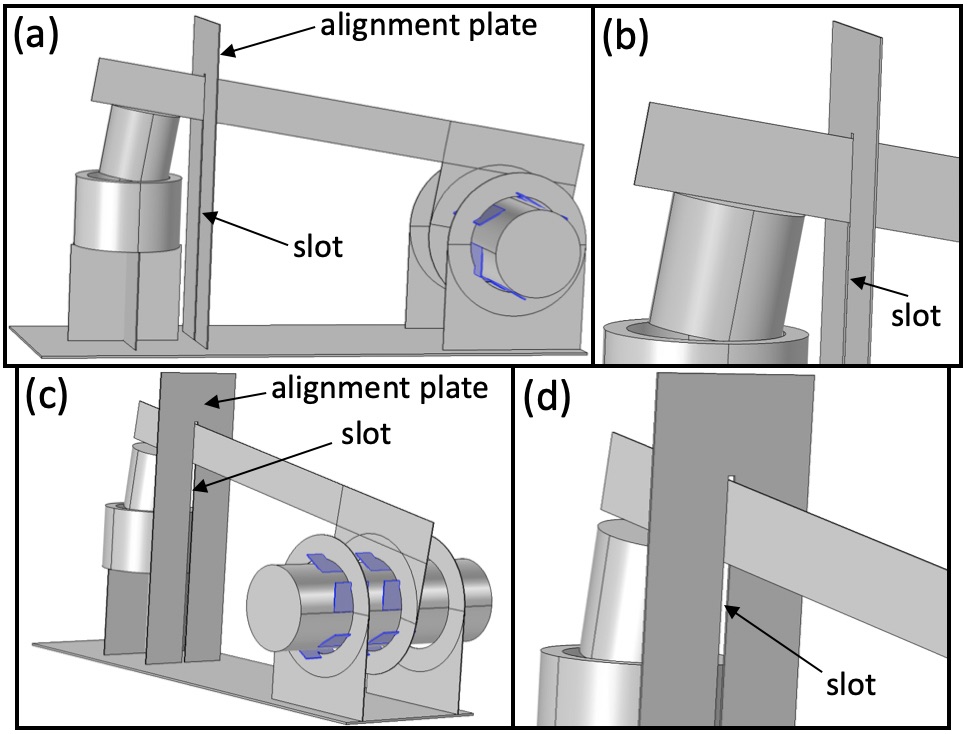,width=3.4in}
\vspace{-1.7em}
\caption{\small{A long narrow slot in the alignment plate keeps the moment arm in a single plane. This slot also restricts the moment arm from rotating more than $\approx 12^\circ$. }}
\vspace{-0.5em}
\label{fig:limiter}
\end{figure}
A 50$\mu$m-thick Al alignment plate, shown in Figure \ref{fig:limiter}, is used to ensure that the moment arm (and hence the input ring) always moves in a plane perpendicular to the shaft. This is accomplished by constraining the moment arm to only move through a narrow slot (100$\mu$m-wide) in the alignment plate. This slot also limits the magnet from moving completely out of the coil. 

\subsection{Starting Torque \& Mechanical Losses} 
Three types of torques need to be overcome in each ratchet to produce motion. One is the friction torque arising from the contact between the elastic beams and the ring. Another is due to the energy dissipated in the deflected elastic beams when they are released after crossing over the peak (in the patterned hole) into the falling edge. And lastly to lift the weight of the magnet against gravity. 

\subsubsection{Friction} 
Each of the elastic beams is $l=0.5$mm long and $w=1$mm wide, hence their bending stiffness is $k=\frac{2.5\text{GPa}}{4} \frac{t^3 w}{l^3}\approx $ 10N/m. When inside the patterned hole, they are pre-deflected (pre-tensioned) by an amount no larger than $\Delta y = $ 0.2mm (estimate), corresponding to a contact force of $F_\text{contact} = $ 2mN per beam. 
Assuming a friction coefficient of $\mu_s=0.1$, this corresponds to a starting torque of $6\cdot \mu_s \cdot F_\text{contact} \cdot r_\text{shaft} = $ 1.2$\mu$Nm per ratchet. 

\subsubsection{Elastic dissipation} 
At any given time only one of the 6 beams is released from the peak into the valley due to the beams not being perfectly 60$^\circ$ apart from each other. Thus in a rotation of $4^\circ \approx $ 0.07rad, an energy of $\frac{1}{2}k{\Delta y}^2 - \frac{1}{2}k(\Delta y-25\mu \text{m})^2 = $ 0.05$\mu$J is dissipated. By energy equivalence, this corresponds to a starting torque of $\frac{0.05\mu \text{J}}{0.07\text{rad}} = $ 0.7$\mu$Nm. 

\subsubsection{Potential energy} 
The 25mg magnet at the end of the 8mm long moment arm, which always remains almost horizontal, exerts a torque of 0.25mN $\cdot$ 8mm = 2$\mu$Nm. 

The total estimated starting torque is the sum of the above three and thus = 3.9$\mu$Nm. 
Experimentally it is found that an applied torque of 4.4$\mu$Nm is sufficient to produce motion. This minimum starting torque needed determines the minimum coil current needed to produce motion. 

Using finite element simulations we find the average magnetic field seen by the coil to be $B_\text{avg} \approx 0.1$T. 
This average is not low because the magnet undergoes only small displacements never being very distant from the coil. The 8mm long moment arm greatly reduces the force the coil needs to generate to produce 4.4$\mu$Nm of torque. $F_\text{coil} $(needed) = 4.4$\mu$Nm/8mm = 0.55mN = $n_\text{turns} \cdot B_\text{avg} \cdot I_\text{coil} \cdot 2\pi r_\text{coil} \Rightarrow I_\text{coil}$ = 0.5mA. The heat loss in the coil at this current value will be $I_\text{coil}^2R_\text{coil} $ = 0.38mW, and the voltage across the coil will be $V_\text{coil}$ = 0.75V. However, the off-the-shelf electronics components used in this paper can't operate below 1V and thus the voltage across the coil would be $V_\text{coil}$(actual) = 1V $\Rightarrow I_\text{coil}$(actual) $\approx$ 0.6mA which is more than the current required to guarantee function. 

\subsection{Power Electronics} 

\begin{figure}[h]
\centering
\epsfig{file=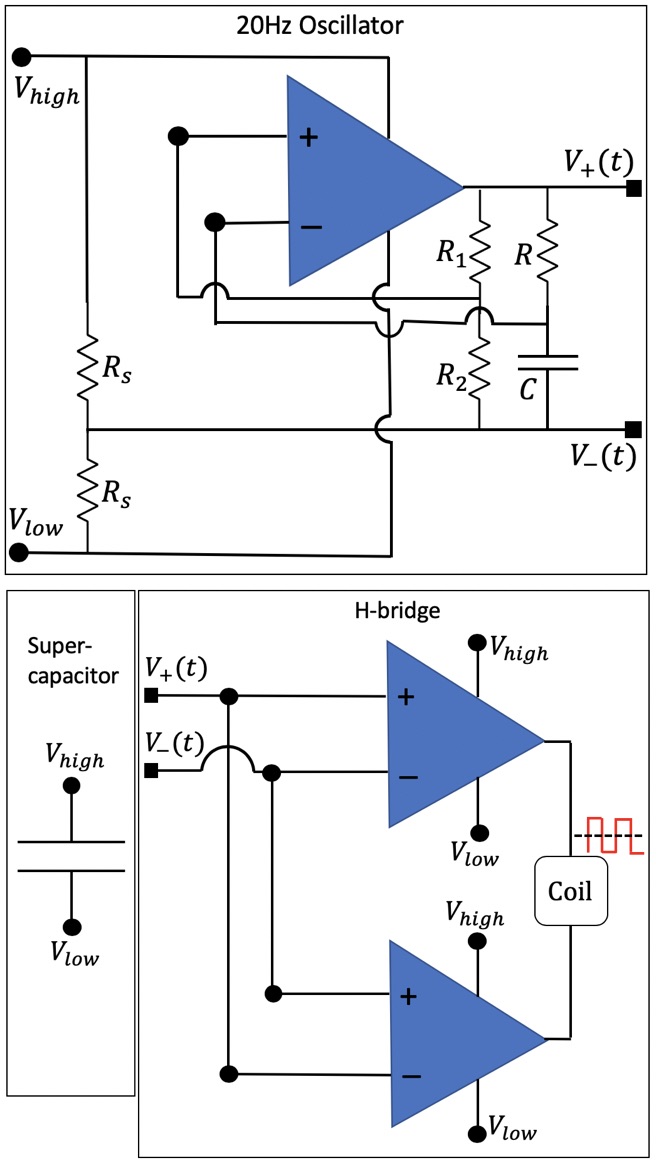,width=2.2in}
\vspace{-0.5em}
\caption{\small{Conceptual circuit diagram. Supercapacitor acts as the supply for the coil, oscillator and the H-bridge. The standard opamp based oscillator circuit functions by charging and discharging the capacitor C, whose time constant is tuned using R. }}
\vspace{-1.5em}
\label{fig:ckt}
\end{figure}

Figure \ref{fig:ckt} shows the schematic of the 3 constituent components of the driving electronics. 
An 11mF supercapacitor from Seiko (CPH3225A) is used as a power source for our device. It has an internal resistance of $160\Omega$ and can be charged up to 3.3V. A resistive divider with $R_s=5.6$k$\Omega$ is used to provide a virtual ground. An opamp based oscillator circuit is used to generate a 20Hz oscillating waveform. Time period of oscillations is given by $T = 2RC\ln(\frac{1+\beta}{1-\beta})$, where $\beta = \frac{R_2}{R_1+R_2}$. Choosing $R_1=R_2=56$k$\Omega$ sets $\beta = 0.5$ and $T \approx 2RC$. Choosing $R=10$k$\Omega$ and $C=2.2\mu$F sets $f=\frac{1}{T}$ near 20Hz. 

This periodic waveform is then fed to an H-bridge made out of 2 opamps which alternates the connection polarity of the coil to the supercapacitor at 20Hz. The supercapacitor discharges through the coil and the opamps stop functioning below a supply voltage of 1V at which point the coil stops receiving alternating supply voltage and the device stops functioning. 

\begin{figure}[h]
\centering
\epsfig{file=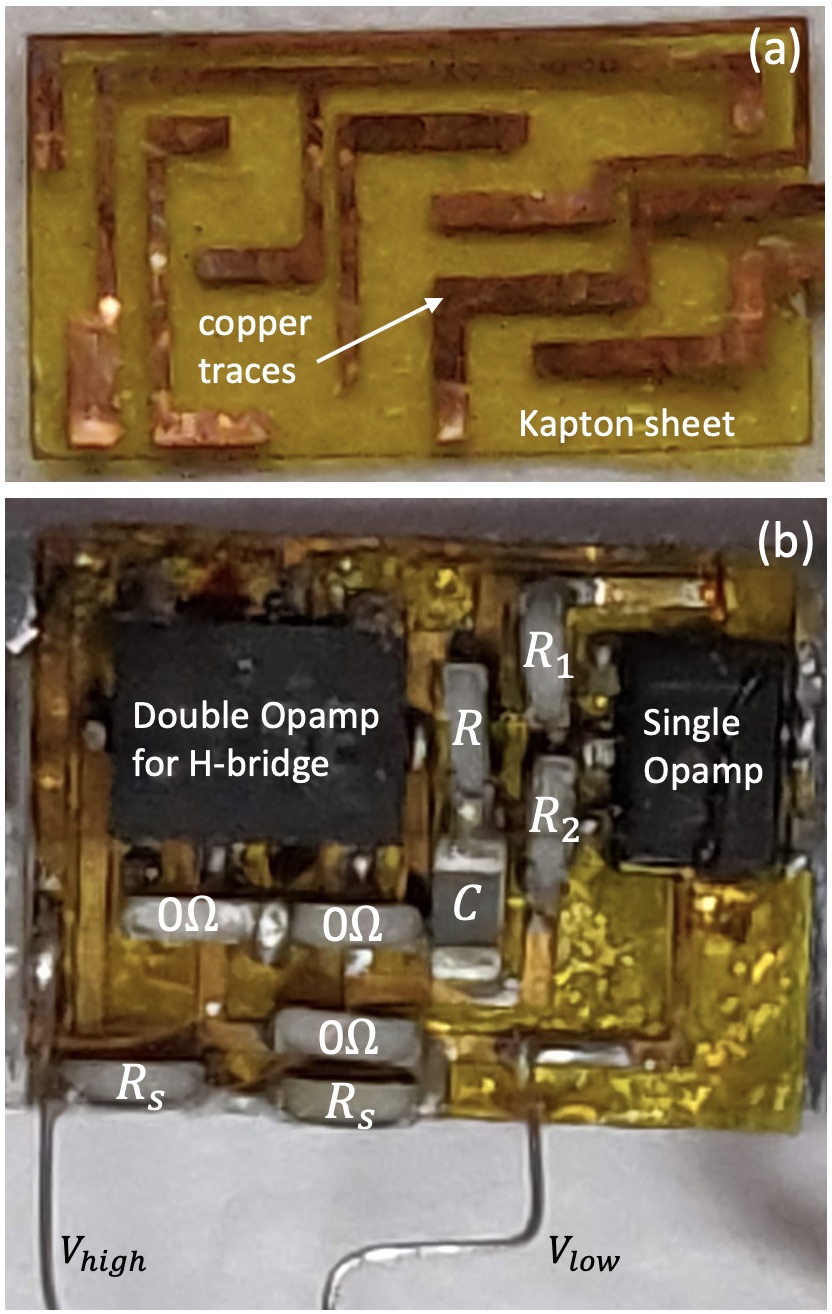,width=2.6in}
\vspace{-0.4em}
\caption{\small{Actual circuit. (a) To avoid any manual alignment copper traces are laser-cut in place, and then the Kapton + adhesive laminate is hot pressed on to it. (b) The surface mount electrical components are glued and soldered in place manually. }}
\vspace{-1.4em}
\label{fig:real-ckt}
\end{figure}
A 12.7$\mu$m-thick Kapton sheet with an 18$\mu$m-thick double-sided adhesive film attached to it acts as the substrate of our circuit. The copper traces acting as wiring are laser cut from a 25$\mu$m thick Copper sheet and then bonded to the substrate using heat and pressure, as seen in Figure \ref{fig:real-ckt}. The surface mount opamps, 0402 resistors, 0603 capacitor and 0402 zero resistance jumpers are glued to the substrate and then soldered to the Copper wiring using solder paste and a hot air gun. Jumpers are used to make electrical connection paths that otherwise intersect with existing copper traces. The double opamp (TLV342 RUG) and the single opamp (TLV341 DRL) units each weight around 2mg and are the heaviest parts in the circuit. 

\subsection{Assembly} 

\begin{figure}[h]
\centering
\epsfig{file=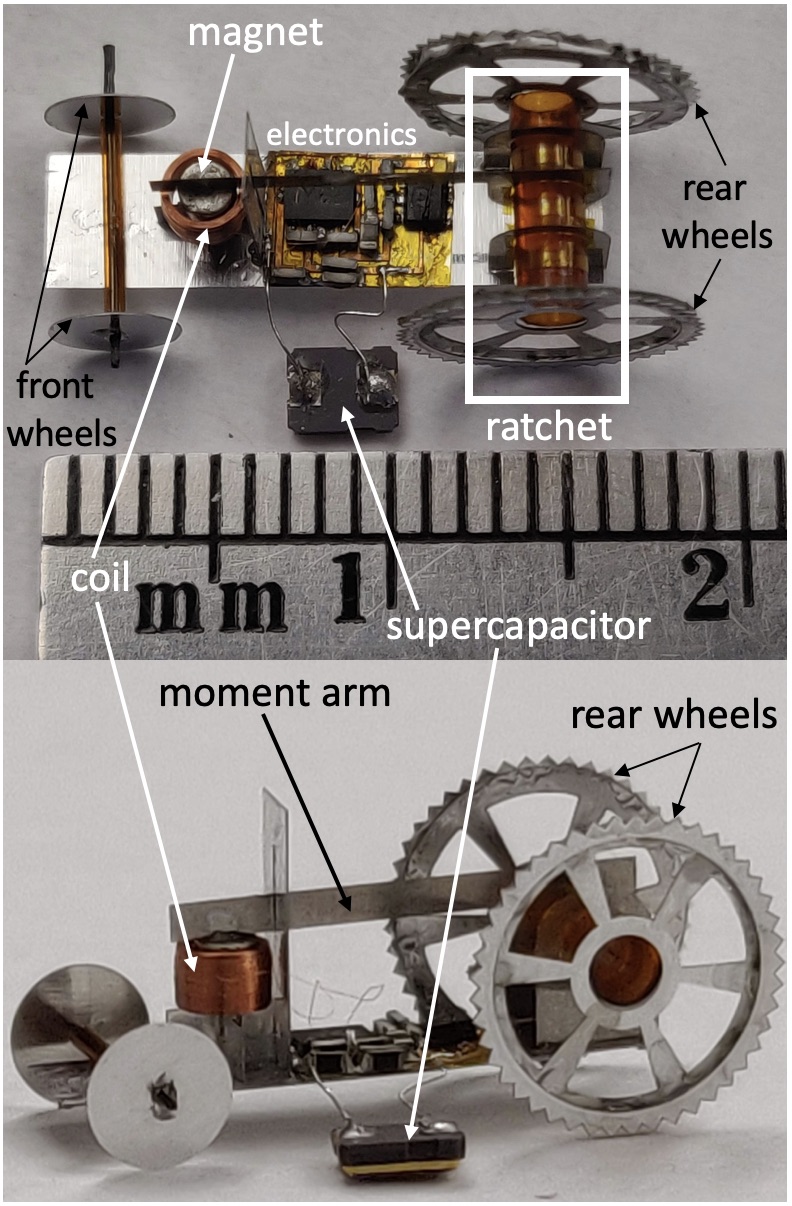,width=2.3in}
\vspace{-0.4em}
\caption{\small{Fully assembled device. The supercapacitor is kept close to the ground so that it can be charged using probes from a function generator and then released quickly. }}
\vspace{-0em}
\label{fig:assembly}
\end{figure}
The completed electronics unit is glued on to the Al base plate in the space below the moment arm as seen in Figure \ref{fig:assembly}. Spiked wheels 8mm in diameter, laser cut from 50$\mu$m-thick Al, are attached to the double-ratchet's shaft. Smaller 3mm diameter wheels, with CF rod as axle and free to rotate inside a Kapton tube, are added to the front to balance the robot. This makes the robot a rear-wheel-drive. The masses of all the constituents after the assembly can be seen in Table \ref{tab:mass}. 

\begin{table}[h]
\vspace{-0.8em}
\normalsize
  \centering 
    \caption{\small Mass distribution of the microbot.}
    \vspace{-0.4em}
    \label{tab:mass}
    \begin{tabular}{|l|r|}
    \toprule 
      \textbf{Sub-component} & \textbf{Mass} \\
          \toprule
      \hline
      \multicolumn{2}{|c|}{Electrical parts}\\ 
      \hline
      Power electronics & 17mg \\
      Coil & 13mg \\
      Magnet + moment arm & 27.3mg \\
      Supercapacitor & 24.1mg \\ 
        \hline
      \multicolumn{2}{|c|}{Structural parts}\\ 
      \hline
      Base plate + Perpendicular supports & 16.2mg \\
      Front wheel assembly & 4.8mg \\
      Rear wheels & 18.7mg \\
      Ratchet tube & 8.6mg \\ 
      \hline
      \bottomrule
      \textbf{Total} & {130mg} \\
      \bottomrule
    \end{tabular}
\vspace{-1.5em}   
\end{table}

\section{Experiments} 

\subsection{Double-ratchet} 

\begin{figure}[h]
\vspace{-0.4em}
\centering
\epsfig{file=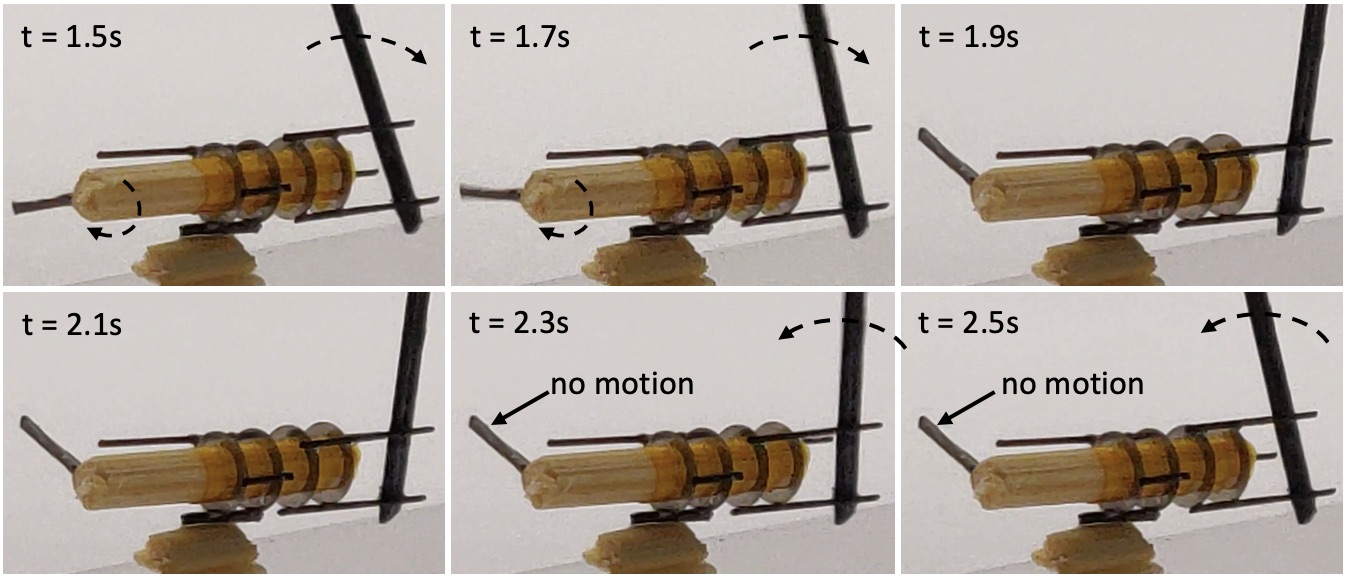,width=3.4in}
\vspace{-1.5em}
\caption{\small{Double-ratchet mechanism operated manually, and thus time stamps are just indicative. Input is provided at the back ratchet, and output is observed using the black CF indicator rod attached perpendicularly to the shaft. Asymmetry in clockwise vs anti-clockwise operation can be observed. }}
\vspace{-1.5em}
\label{fig:ratchet-experiment}
\end{figure}
We verify the functioning of the double-ratchet by providing it a periodic input manually as shown in Figure \ref{fig:ratchet-experiment}. The shaft is engaged to the back ratchet when the input is driven clockwise. This can be seen by noticing the motion of the black indicator attached to the shaft. The second row shows the shaft being grounded, with no motion of the indicator, when the input is driven anti-clockwise. 

\subsection{Rolling using photovoltaics} 

\begin{figure}[h]
\centering
\epsfig{file=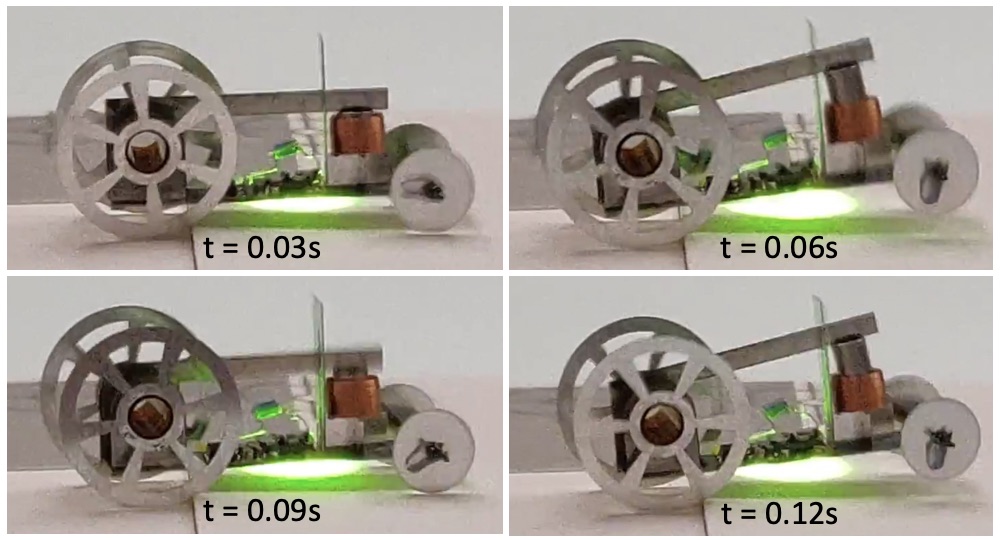,width=3.4in}
\vspace{-1.7em}
\caption{\small{Stationary laser-powered bot with continuously rotating but slipping wheels. The wheels are made to slip by smoothening it out (eliminating the spikes) and then placing them at a bump that they cannot climb due to low traction. }}
\vspace{-1.4em}
\label{fig:pv-bot}
\end{figure}
Before trying the supercapacitor powered version of the bot we tried an alternate power source which is a 1mm$\times$1mm infrared PV cell (MH GoPower 5S0101.4-W) that produces current when a 976nm wavelength laser light (MH GoPower LSM-010) is shone on it. The laser intensity is increased until the PV cell outputs $\approx$1.5V while driving a 1.5k$\Omega$ load. The robot's operation was intermittent since the onboard PV cell moves out of the laser spot (seen as the green spot on the infrared indicator card in Figure \ref{fig:pv-bot}) as soon as the robot rolls forward, and then needs to be repointed which is done manually. So to test the operation we allow the smoothed out rear wheels of the robot to slip in a gap/valley between two cards so that its wheels rotate but the robot doesn't move forward and its PV cell remains in the laser spot. Because of the absence of the spikes on the rear wheels and the heavy supercapacitor, this version of the robot weights 96mg. Note that even while in motion if the laser is somehow shone continuously on the PV cell then a continuous forward motion would be expected. 

\begin{figure}[h]
\vspace{-0.6em}
\centering
\epsfig{file=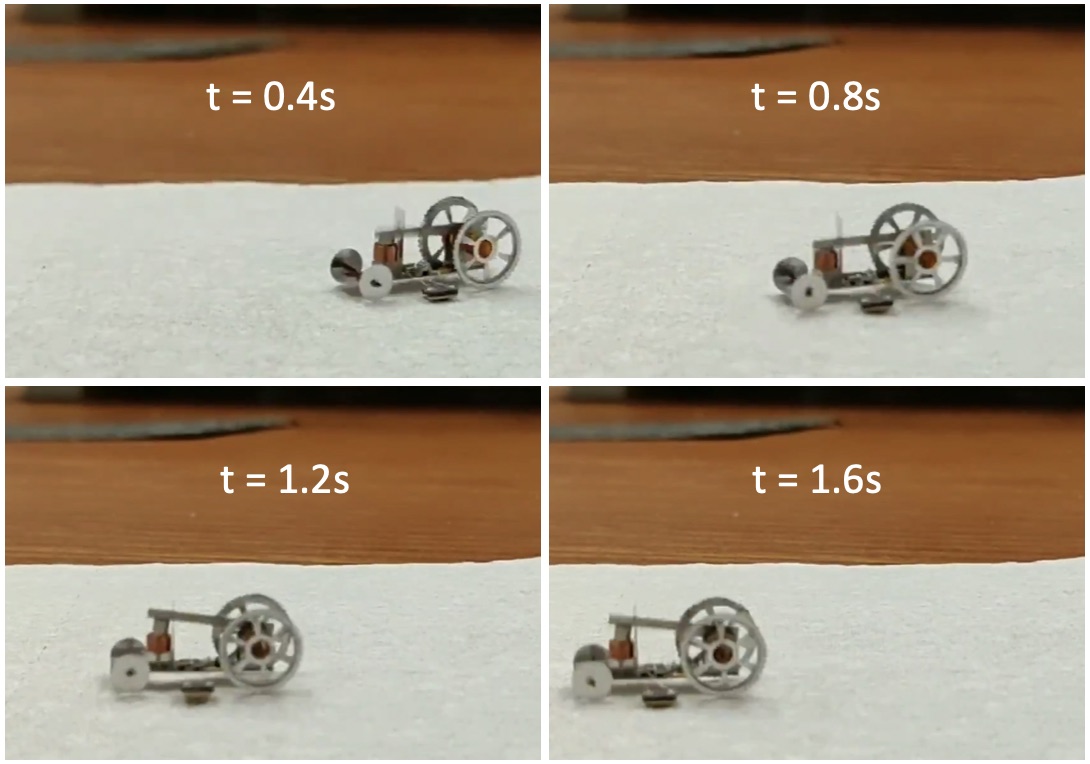,width=3.4in}
\vspace{-1.6em}
\caption{\small{Microrobot rolling forwards in real time. The bot is operated over a piece of paper for better traction and to avoid any slipping between the spiked wheels and the level surface. }}
\vspace{-0.5em}
\label{fig:roll-experiment}
\end{figure}

\subsection{Rolling using supercapacitor} 

The supercapacitor is charged up to 3V (in 1 minute) using an external function generator. After this charge, the voltage across the supercapacitor drops from 3V to 1V in 8s due to it getting discharged via the coil. During this phase, the magnet drives the input of the ratchet at 20Hz resulting in the rear wheels rotating at 300$^\circ$/s and the robot rolling forwards at 27mm/s as seen in Figure \ref{fig:roll-experiment}. If we had a constant 1V battery, then the robot can be kept operating while consuming 0.6mW of power. But since we don't have a constant voltage battery, the average power consumed in the 8s is greater at 2.5mW since the capacitor starts with a higher voltage. 

\section{Conclusions and Future work} 
The wheels and the supporting structures in the robot weight 40mg and can be made much lighter by using carbon fiber or using the material more sparsely. The bot could be made to consume an even lower power if the electronics could function below 1V, but we didn't find any lower voltage light-weight opamps. 

The mechanical work done by the actuator to overcome the mechanical losses in the mechanism is negligible compared to the Joule heat loss in the coil. This Joule heat loss is independent of the actuator frequency. Thus, the wheels can be made to rotate much faster simply by increasing the operating frequency of the actuator, and still consume almost the same amount of power while rolling forwards much faster. 

The proposed double-ratchet can work with any other actuator and convert small periodic motions to continuous rotation. 
Using the same principles, one can make a much smaller rolling robot as well but we expect the availability of off-the-shelf power electronics components to be very restricting at smaller scales, and custom chips would have to be made.

\section*{Acknowledgement}

The authors are grateful to get support from Commission on Higher Education (award \#IIID-2016-005)
and DOD ONR Office of Naval Research (award \#N00014-16-1-2206). 
We would also like to thank Prof. Ronald Fearing for his help and insightful discussions.


\begin{thebibliography}{99}

\bibitem{wood_liftoff} K. Ma, P. Chirarattanon, S. Fuller, and R.J. Wood, ``Controlled Flight of a Biologically Inspired, Insect-Scale Robot,'' {Science}, vol. 340, pp. 603-607, 2013. 

\bibitem{zhang16} Y. Zou, W. Zhang, and Z. Zhang, ``Liftoff of an Electromagnetically Driven Insect-Inspired Flapping-Wing Robot,'' {IEEE Transactions on Robotics}, vol. 32, no. 5, October 2016. 

\bibitem{baybug18} P. Bhushan and C.J. Tomlin, ``Milligram-scale Micro Aerial Vehicle Design for Low-voltage Operation,'' {IROS}, Madrid, Spain, Oct. 2018.  

\bibitem{baybug19} P. Bhushan and C.J. Tomlin, ``Design of the first sub-milligram flapping wing aerial vehicle,'' {MEMS}, Seoul, South Korea, Jan. 2019.  

\bibitem{robofly18} J. James, V. Iyer, Y. Chukewad, S. Gollakota, and S.B. Fuller, ``'Liftoff of a 190 mg Laser-Powered Aerial Vehicle: The Lightest Untethered Robot to Fly,'' {IEEE Int. Conf. on Robotics and Automation}, Brisbane, Australia, May 2018. 

\bibitem{actuator_selection} M. Karpelson, G-Y. Wei, and R.J. Wood, ``A Review of Actuation and Power Electronics Options for Flapping-Wing Robotic Insects,'' {IEEE Int. Conf. on Robotics and Automation}, Pasadena, CA, May 2008. 

\bibitem{pister_10mg} S. Hollar, A. Flynn, C. Bellew, and K.S.J. Pister, ``Solar powered 10 mg silicon robot,'' {MEMS}, Kyoto, Japan, 2003.

\bibitem{contreras17} D.S. Contreras, D.S. Drew, and K.S.J. Pister, ``First steps of a millimeter-scale walking silicon robot,'' {19th International Conference on Solid-State Sensors, Actuators and Microsystems}, 910-913, 2017.

\bibitem{saito16} K. Saito, K. Iwata, Y. Ishihara, K. Sugita, M. Takato, and F. Uchikoba, ``Miniaturized Rotary Actuators Using Shape Memory Alloy for Insect-Type MEMS Microrobot,'' {Micromachines}, 7(4): 58, 2016.

\bibitem{bergbreiter_1mg} R.S. Pierre, W. Gosrich, and S. Bergbreiter, ``A 3D-printed 1 mg legged microrobot running at 15 body lengths per second,'' {Hilton Head Solid-State Sensors, Actuators, and Microsystems Workshop}, Hilton Head Island, SC, June 2018.

\bibitem{rus06} B.R. Donald, C.G. Levey, C.D. McGray, I. Paprotny, and D. Rus, ``An untethered, electrostatic, globally controllable MEMS micro-robot,'' {Journal of Microelectromechanical Systems}, Vol. 15, No. 1, pp 1-15, 2006.

\bibitem{lin_crawling} M. Qi, Y. Zhu, Z. Liu, X. Zhang, X. Yan, and L. Lin, ``A fast-moving electrostatic crawling insect,'' {MEMS}, Las Vegas, NV, Jan. 2017.

\bibitem{inchworm02} R. Yeh, S. Hollar, and K.S.J. Pister, ``Single mask, large force, and large displacement electrostatic linear inchworm motors,'' {Journal of Microelectromechanical Systems}, Vol. 11, No. 4, pp 330-336, 2002.

\bibitem{inchworm12} I. Penskiy and S. Bergbreiter, ``Optimized electrostatic inchworm motors using a flexible driving arm,'' {Journal of Micromechanics and Microengineering}, Vol. 23, No. 1, pp 1-12, 2012. 

\end{thebibliography}
\end{document}